\documentclass[runningheads]{llncs}
\usepackage{graphicx}
\usepackage{tikz}
\usepackage{comment}
\usepackage{subcaption}
\usepackage{amsmath,amssymb} % define this before the line numbering.
\usepackage{color}
\usepackage{multirow}
\usepackage{array}
\usepackage{stfloats}
\usepackage[pagebackref,breaklinks,colorlinks]{hyperref} 

\usepackage[accsupp]{axessibility}  % Improves PDF readability for those with disabilities.

\usepackage[width=122mm,left=12mm,paperwidth=146mm,height=193mm,top=12mm,paperheight=217mm]{geometry}

\newcommand{\numlabels}{200}
\newcommand{\OURS}{ScanNet200}

\newcolumntype{?}{!{\vrule width 1.2pt}}

\begin{document}
% \linenumbers
\pagestyle{headings}
\mainmatter
\def\ECCVSubNumber{6493}  % Insert your submission number here

\title{Language-Grounded Indoor 3D Semantic Segmentation in the Wild} 

\begin{comment}
\titlerunning{ECCV-22 submission ID \ECCVSubNumber} 
\authorrunning{ECCV-22 submission ID \ECCVSubNumber} 
\author{Anonymous ECCV submission}
\institute{Paper ID \ECCVSubNumber}
\end{comment}
%******************

%\begin{comment}
\titlerunning{}
\author{David Rozenberszki\inst{1} \and
Or Litany\inst{2} \and
Angela Dai\inst{1}}
\authorrunning{D. Rozenberszki et al.}
\newcommand{\samelineand}{\qquad}
\institute{\textsuperscript{1}Technical University of Munich \qquad \textsuperscript{2}NVIDIA \\ 
\begin{center}
\href{https://rozdavid.github.io/scannet200}{https://rozdavid.github.io/scannet200}
\end{center}}

%\end{comment}
%******************

\maketitle

\begin{figure}[!ht]
    \centering
    \includegraphics[width=\textwidth,keepaspectratio]{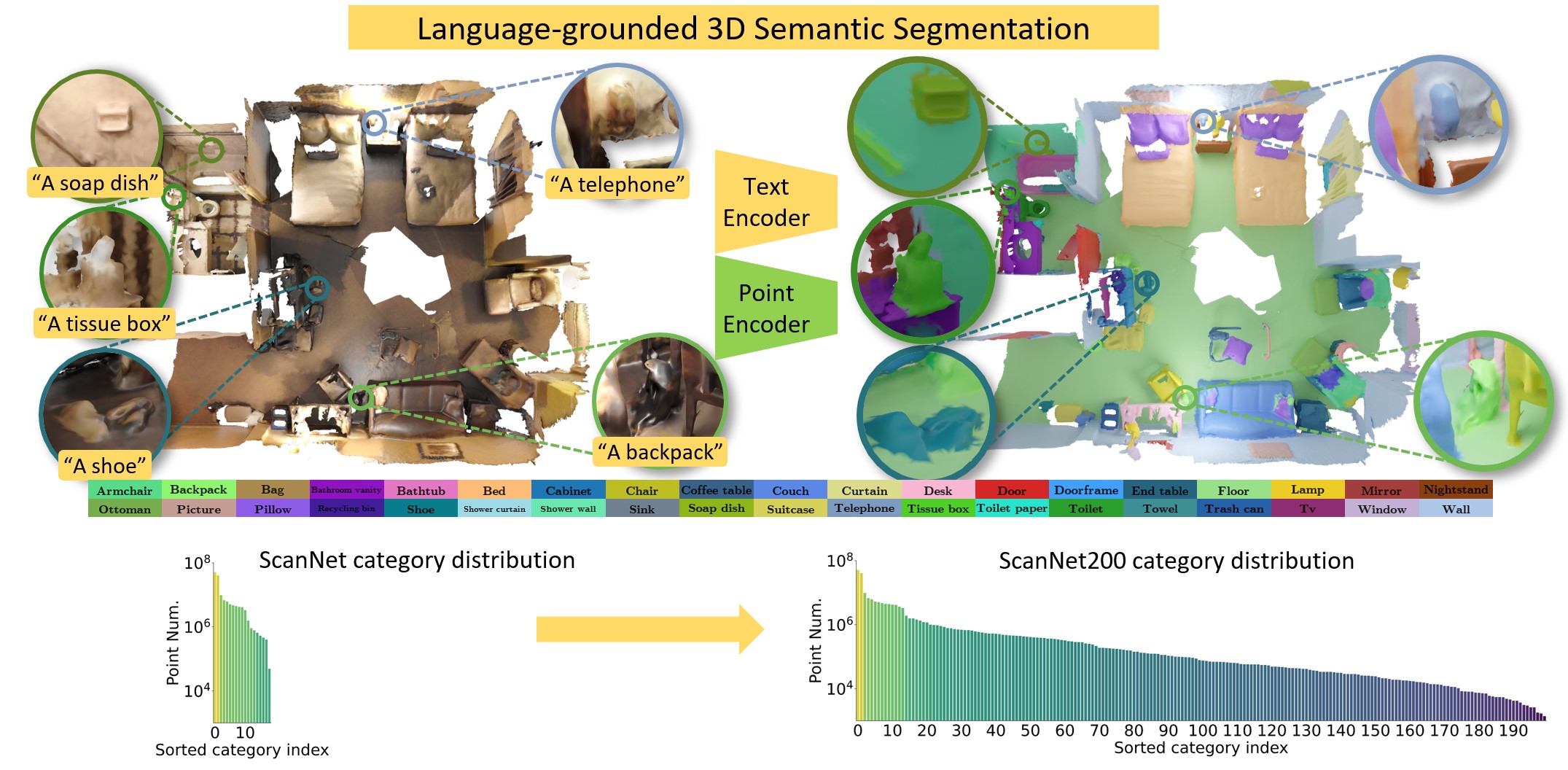}
    \caption{
    We present the \OURS{} benchmark, which studies \numlabels{}-class 3D semantic segmentation -- an order of magnitude more categories than previous 3D scene understanding benchmarks.
    To address this challenging task, we propose to guide  3D feature learning by anchoring it to the richly-structured text embedding space of CLIP for the semantic class labels.
    This results in improved 3D semantic segmentation across the large set of class categories.} 
    \label{fig:teaser}
\end{figure}

\begin{abstract}
Recent advances in 3D semantic segmentation with deep neural networks have shown remarkable success, with rapid performance increase on available datasets.
However, current 3D semantic segmentation benchmarks contain only a small number of categories -- less than 30 for ScanNet and SemanticKITTI, for instance, which are not enough to reflect the diversity of real environments (e.g., semantic image understanding covers hundreds to thousands of classes). 
Thus, we propose to study a larger vocabulary for 3D semantic segmentation with a new extended benchmark on ScanNet data with 200 class categories, an order of magnitude more than previously studied.
This large number of class categories also induces a large natural class imbalance, both of which are challenging for existing 3D semantic segmentation methods.
To learn more robust 3D features in this context, we propose a language-driven pre-training method to encourage learned 3D features that might have limited training examples to lie close to their pre-trained text embeddings.
Extensive experiments show that our approach consistently outperforms state-of-the-art 3D pre-training for 3D semantic segmentation on our proposed benchmark (+9\% relative mIoU), including limited-data scenarios with +25\% relative mIoU using only 5\% annotations. 

 \\

%\textbf{Project Page:} \href{https://rozdavid.github.io/scannet200}{https://rozdavid.github.io/scannet200} %\\
%
%\textbf{Implementation:} \href{https://github.com/RozDavid/LanguageGroundedSemseg} {https://github.com/RozDavid/LanguageGroundedSemseg}
\end{abstract}
\section{Introduction}

In recent years, remarkable advances have been made in 3D semantic segmentation as a core task underlying 3D perception for myriad applications, including robotics, autonomous navigation, and mixed reality.
The introduction of several large-scale real-world 3D datasets \cite{scannet,matterport3d,semantickitti} has led to rapid developments in data-driven 3D deep learning techniques, with an emphasis on point- and sparse-voxel-based approaches \cite{submanifold,kpconv,minkowski,xu2021rpvnet,occuseg}.
However, popular benchmarks such as ScanNet~\cite{scannet} or SemanticKITTI~\cite{semantickitti} focus on a limited number of class categories (20 and 28 classes, respectively), and thus these label sets do not well-represent the diversity and complexity of real scene content that would be encountered in the wild.
In contrast, common image segmentation benchmarks \cite{pascal,lin2014microsoft} contain over 80 annotated class labels, with recent large-vocabulary image challenges~\cite{lvis} presenting over 1000 categories for recognition tasks. 

Thus, we propose to address a larger-vocabulary setting for 3D semantic segmentation.
In particular, we focus on the indoor domain and consider 3D scans of ScanNet~\cite{scannet} where a variety of different object categories are seen in the RGB-D scans despite its benchmark evaluating on only 20 classes.
We present \OURS{}, a \numlabels-class 3D semantic segmentation benchmark, considering an order of magnitude more class annotations than previously considered. 
This new set of classes includes both finer-grained categories of previous classes as well as a large number of previously unaddressed classes.
This induces a much more challenging setting reflecting the naturally observed semantic classes already seen in the raw ScanNet RGB-D observations, where the data also reflects naturally encountered class imbalances (e.g., walls and floors are seen much more often than nightstands, which are also seen far more often than fire extinguishers).
In addition considering the setting where all dense annotations are available for train scenes for the \numlabels{} classes, we also consider limited annotation scenarios with only sparse annotations per scene, given the expense of 3D data annotation.

In order to address this challenging new benchmark for 3D semantic segmentation, we explore standard techniques for data and loss balancing for the much larger number of class categories. 
In combination with the most effective techniques, we further observe that, unlike the limited, imbalanced geometric content, state-of-the-art language models have observed and attained rich representations of all categories, and so can induce a better structure onto learned 3D embeddings.
Thus, we propose to ground 3D feature learning with strong pre-trained CLIP text features to construct a richly-structured 3D feature representation space.
To this end, we formulate a language-grounded pre-training by mapping learned 3D features to pre-trained language embeddings with a contrastive loss.
This enables a more robust 3D representation learning under imbalanced and limited 3D observations.
Experiments on our \OURS{} semantic segmentation as well as semantic segmentation in the limited data regime demonstrate the effectiveness of our language-grounded 3D semantic segmentation.

\noindent
In summary, our contributions are:
\begin{itemize}
    \item We propose  a new \numlabels-class 3D semantic segmentation benchmark on real-world 3D ScanNet scene data, considering an order of magnitude more category annotation labels than existing 3D semantic segmentation benchmarks.
    \item In order to guide the construction of robust 3D semantic feature representations for this challenging task, we propose to align geometric feature extraction to the category embedding of the CLIP pretrained language model. This results in improved performance both overall and in the rarely seen, including in the limited-data regime.
\end{itemize}
\section{Related Work}

\subsubsection{3D Semantic Segmentation.}
With the introduction of large-scale annotated real-world 3D datasets \cite{scannet,matterport3d,semantickitti}, 3D semantic segmentation has seen significant focus in recent years with various deep learning-based methods developed around different 3D representations.
Early works tackled 3D semantic segmentation on dense volumetric grids \cite{scannet,dai20183dmv}, but were limited in cubic growth in memory and compute.
The introduction of PointNet~\cite{qi2017pointnet} presented a point-based alternative with strong memory efficiency by operating on unstructured point clouds, with various methods introducing local operators to better learn neighborhood structures \cite{qi2017pointnet++,kpconv,xu2018spidercnn}.
Hierarchical grid structures such as octrees provided a more structured alternative for grid-based reasoning without dense memory consumption \cite{riegler2017octnet}.
Recently, the introduction of sparse 3D convolutions \cite{submanifold,minkowski} enabled significant performance improvements by leveraging a structured space representation in a sparse fashion to operate efficiently at high resolutions.
In this work, we also adopt a sparse 3D convolutional backbone to explore language-guided pre-training for larger-vocabulary semantic segmentation.

\subsubsection{3D Representation Learning.}
Inspired by the success of contrastive frameworks for 2D image representation learning \cite{van2018representation,chen2020simple,momentum_contrast,simsiam}, 3D representation learning has begun to see exploration in unsupervised contrastive pre-training.
PointContrast~\cite{pointcontrast} demonstrated the effectiveness of unsupervised contrastive pre-training for 3D scene understanding tasks, with various methods introducing augmentation alternatives for 3D pre-training \cite{rao2021randomrooms,huang2021spatio,chen20214dcontrast,zhang_depth_contrast}.
Contrastive Scene Contexts~\cite{scene_contrast} introduced an unsupervised contrastive pre-training in the context of data-efficient 3D scene understanding with limited reconstruction and limited annotations available. 
In contrast to these 3D pre-training methods, we propose a supervised multi-modal 3D representation learning guided by text encoded features to learn a more robust feature representation space covering significantly more class categories than previously studied for 3D.
Inspired by the data-efficient scene understanding of \cite{scene_contrast}, we additionally study a limited annotations scenario for our \OURS{} benchmark. 

Additionally, Mix3D~\cite{mix3d} presented a data augmentation scheme to mix multiple 3D scenes together to generate semantic segmentation that is more robust against undesired context biases.
Our instance-based sampling when fine-tuning the learned language-guided 3D features is inspired by the scene mixing, but operates at an instance level to help mitigate class imbalances.
Previous methods have also leveraged text embeddings in 3D learning for zero-shot pointcloud segmentation ~\cite{DBLP:journals/corr/abs-2108-06230,DBLP:journals/corr/abs-2104-04980} and classification~\cite{pointclip}. More recently, CLIP~\cite{clip} was shown as a powerful conditioner for generative 3D models \cite{clipforge,text2mesh}.
We also aim to leverage powerful CLIP text embeddings for robust 3D semantic pre-training.

\subsubsection{3D Scene Understanding Benchmarks.}
Recently, various large-scale real-world 3D scene understanding benchmarks have been established.
Early benchmarks such as the NYUv2 dataset~\cite{silberman2012indoor} introduced RGB-D frame-based annotations on a limited number of frames (e.g., 1449 for NYUv2).
ScanNet~\cite{scannet} presented a much larger-scale RGB-D dataset and benchmark with 1513 densely-annotated reconstructed 3D scans.
While it contains hundreds of raw annotated label data, the ScanNet benchmark evaluates only 20 class categories for its 3D scene understanding tasks.
Similarly, Matterport3D~\cite{matterport3d} presents a large-scale RGB-D dataset with a 20-class semantic segmentation evaluation.
Additionally, SemanticKITTI~\cite{semantickitti} established an outdoor 3D dataset and benchmark for LiDAR-based scene understanding with 28 class category annotations.
We present our \OURS{} benchmark based on ScanNet scene data with an order of magnitude more classes than previous benchmarks.

\subsubsection{Class Imbalance.}
Real-world dataset annotations tend to contain natural class imbalances which can lead to skewed learning of more frequently observed class categories.
Despite the lack of study on mitigating class imbalances in 3D, various methods have been presented to address them in 2D image understanding tasks.

In particular, class imbalance in image classification problems is often addressed by oversampling underrepresented categories with strong data augmentation techniques to obtain an evenly-distributed dataset.
Various methods have been introduced towards data-sampling-based re-balancing, for instance random oversampling of underrepresented classes \cite{random_oversampling,gtpaste-Yan2018SECONDSE,wang2021pointaugmenting}, sampling novel poses of known categories \cite{manhardt2019roi}, undersampling overrepresented classes \cite{undersampling}, frequency-based sampling \cite{taunorm}, as well as feature-based or generative sampling \cite{gen_samples,trainable_undersampling,optimal_transport}. 
%The most similar to our methods \cite{gtpaste-Yan2018SECONDSE,wang2021pointaugmenting} are Lidar point cloud based sampling strategies focusing on pasting instances from vocabulary and filtering original Lidar points by occlusion. 
Inspired by such approaches, we propose a 3D instance-based sampling to mitigate class imbalances for 3D semantic segmentation.

Alternative methods have been proposed to re-balance the loss for image understanding tasks \cite{groupedsoftmax,droploss,focalloss}.
In particular, the focal loss~\cite{focalloss} has been shown to be effective for 2D object detection and semantic segmentation by focusing the training on hard examples or to instance contours \cite{biasutti2019lu}. We also study the effect of focal loss balancing for the 3D semantic segmentation task.

\section{Method}

Our approach tackles the \numlabels{}-class 3D semantic segmentation task on ScanNet~\cite{scannet} data, exploiting well-structured language models that have trained on rich observations across all category labels. 
In particular, we leverage pre-trained text embeddings from CLIP~\cite{clip} as anchors to which we learn to map geometric features during the pre-training step.
We then use these language-grounded features for fine-tuning downstream 3D semantic segmentation.
During fine-tuning, we further address the class imbalance by instance-based augmentation as well as focal loss-based class-balancing for the downstream loss.

\subsection{Language-Grounded 3D Feature Learning}\label{sec:clip_pretraining}

\begin{figure}[!t]
    \centering
    \includegraphics[width=\textwidth,keepaspectratio]{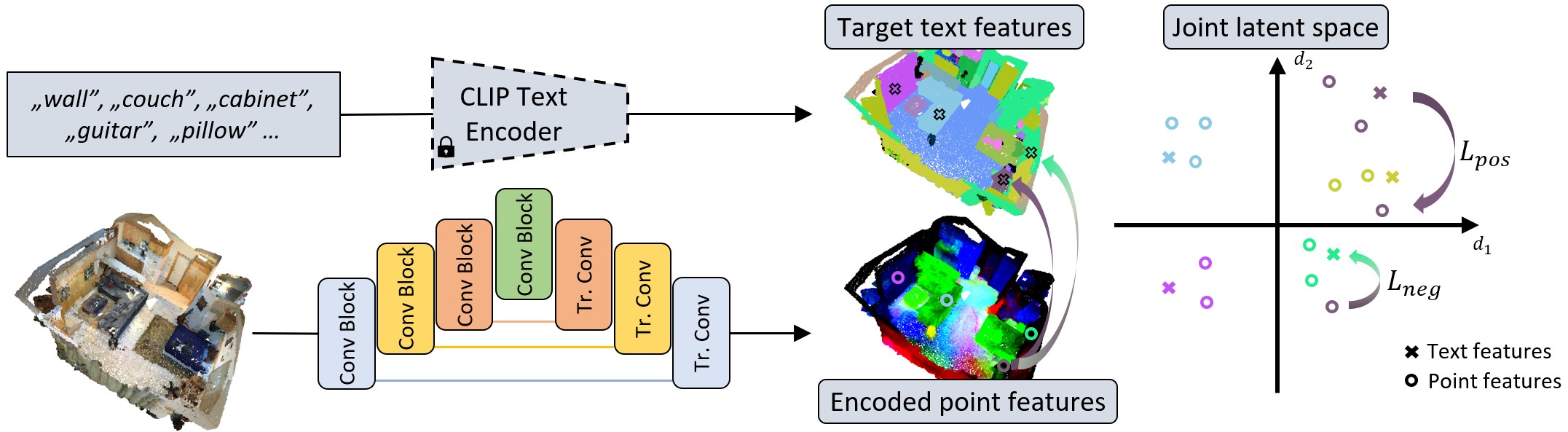}
    \caption[]{
    During pre-training, we guide 3D feature learning by mapping learned features to text encoded anchors of the corresponding semantic labels, constructed by a constrastive loss between text and 3D.
    This establishes a more robust  3D feature representation space guided by the rich structure of the text embeddings.} 
    \label{fig:contrastive_pipeline}
\end{figure}

As training data for language-based models are available in far greater quantities than 3D semantic annotations, we propose to ground 3D feature learning to well-structured, pre-trained text encodings.
This enables a more robust construction of a learned feature space guided towards a highly-structured, rich text feature space, to support downstream 3D semantic segmentation.
An overview of our language-grounded 3D pre-training is shown in Figure~\ref{fig:contrastive_pipeline}.

\subsubsection{Text Encoder.} 
We leverage a pre-trained CLIP~\cite{clip} to map semantic labels to text features.
Note that our approach is agnostic to the specific language model used, but we found CLIP's multi-modal training is well-suited to our language-3D pre-training.
We refer to the supplemental for additional analysis on alternative text embeddings.

During pre-training, the text encoder is kept fixed, and takes the $N_{\textrm{class}}=$\numlabels{} target semantic labels in their text form, tokenizes them, and encodes them to their text encodings to $f^t_1,...,f^t_{N_{\textrm{class}}} \in \mathbb{R}^D$, where $D$ is the dimensionality of the text representation space. 
We leverage the text features $f^t_i$ to anchor learning of 3D features such that learned 3D features will lie close to text encodings if they represent the same semantic class.

\subsubsection{3D Encoder.} 
For 3D feature extraction, we employ a state-of-the-art sparse 3D convolutional U-Net \cite{minkowski}.
Our 3D encoder backbone takes as input a sparse voxelized 3D scan $\mathcal{S}$, with RGB color as input features, and produces for each sparse voxel location a 3D feature $f^s_i\in \mathbb{R}^D$.

\subsubsection{Text-supervised Contrastive optimization.}
We then train the 3D feature encoder to map to the well-structured space of the text model by formulating a contrastive objective to bring together the different data modalities.
For a 3D scan $\mathcal{S}$ with all $N_p$ sparse voxel locations in the current batch, we map together 3D features $f^s_i$ to text features $f^t_{h(i)}$ representing the semantic label text:
\begin{equation}
    \mathcal{L}_{pos} = \sum_{i=1}^{N_p} max\left(0, \frac{f^s_i \cdot  f^t_{h(i)}}{|f^s_i| \cdot |f^t_{h(i)}|} - t_{pos}\right),
\end{equation}
where $h(i)$ is the semantic text label for location $i$, and $t_{pos}$ is a threshold value for gradient clipping.

Similarly, multiple non-matching semantic text features, sampled from all text semantic labels, are pushed away from the learned features as negatives:
\begin{equation}
    \mathcal{L}_{neg} = \sum_{i=1}^{N_p} \frac{1}{|M|}  \sum_{j\in M} max\left(0, t_{neg} -\frac{f^s_i \cdot f^t_j}{|f^s_i| \cdot |f^t_j|}\right),
\end{equation}
where $M \in N_{\textrm{class}}$ are a set of semantic label encodings different from $i$, $f^t_j$ is the corresponding text feature, and $t_{neg}$ is a threshold value for gradient clipping.

We found that a cosine distance between features empirically produced the best results compared to alternative distance measures such as $\ell_1$, $\ell_2$, or MSE.
This allows for more flexibility in the feature learning by constraining only vector directions, and is similarly reflected in CLIP-driven image classification \cite{lseg,open_vocab}. 

The final language-3D pre-training objective is then:
\begin{equation}
 \mathcal{L} = \mathcal{L}_{pos} + \lambda \mathcal{L}_{neg}
\end{equation}
where $\lambda$ weights the effect of the multiple negatives with the positive loss. 
We found empirically that negative sampling was necessary for effective 3D representation learning, rather than employing positive text associations only.  During optimization, multiple possible point feature trajectories are converging to the target anchors, and we encourage the solutions that maximize cluster separation at all times (see Sec.~\ref{sec:results} for additional analysis). Additionally, as we sample target feature anchors from the complete set of categories, we are able to maximize cluster separation within categories rarely appearing together in the same scenes, in contrast to unsupervised algorithms.

%While this contrastive pre-training imbues some knowledge of semantic labels due to the joint embedding with text anchors, it only optimizes the representation space rather than an output probability distribution over the possible classes, so we leverage this as a pre-training step for further downstream fine-tuning.

\subsection{3D Semantic Segmentation Fine-tuning}

We use the language-grounded pre-trained 3D features for fine-tuning for 3D semantic segmentation. 
Here, we also directly address the inherent class imbalance due to the natural long-tail distribution of the class categories in densely-annotated 3D scenes (e.g., far more walls and floors than lamps or dumbbells).
In particular, we address this through data augmentation for class balancing as well as a class-balanced loss.

\subsubsection{Class re-balancing by instance sampling.}\label{sec:instance_sampling}

\begin{figure}[t]
 \centering
    \includegraphics[width=\textwidth,keepaspectratio]{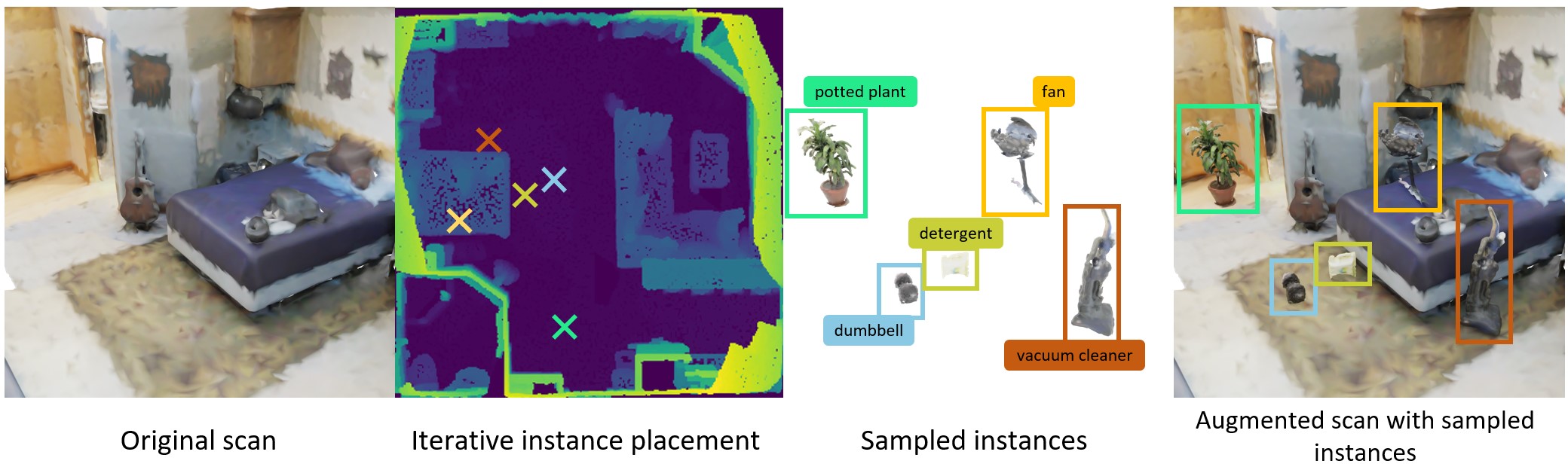}
    % \vspace{-0.5cm}
    \caption{
    Our instance sampling augments scenes during training with by placing rarely-seen class category instances into them, breaking unduly specific context dependencies that can be easily learned from only a few examples.}
    \label{fig:instance_placement}
\end{figure}
We observe that since rare classes are not only infrequently observed but are often small objects and thus represented by smaller sets of points or voxels, they often overfit to recognizing both the surrounding context and the object.
We thus propose to augment scenes by placing instances of infrequently seen class categories in them and breaking overly specific context dependencies for recognition.

An overview of our instance sampling is shown in Figure~\ref{fig:instance_placement}.
We obtain instances from \OURS{} semantic instance annotations, and sample from instances of rare class categories from train scenes. We note here, that we relied on the available ScanNet instance annotations, but since we are augmenting long tail categories only, sparsely appearing in all scenes, the conversion from semantic to instance segmentations comes essentially free with surface label clustering. 
We place these sampled instances in potentially physically valid locations in a new scene.
To this end, we compute a height map of the scene in which the object is to be inserted and iteratively sample instance centroid candidates where the new object can be placed. 
Any sampled object center where the inserted object would collide with existing objects, based on bounding box overlap, is discarded. For all accepted placements we update the height map and continue with the iterations until the condition on the number of samples is met.
This enables class re-balancing by over-sampling rare categories and breaking unduly specific context dependencies for recognition. For additional implementation details please refer to Section 8 in our supplemental material.

\subsubsection{Class-balanced loss.}\label{sec:loss_functions}
As instance sampling-based data augmentation will not fully balance classes (e.g., walls, floors, and other frequently seen categories still dominate), we also consider the class balancing of the loss function.
Rather than a standard cross entropy loss for semantic segmentation, we adapt a focal loss~\cite{focalloss} which was shown to be effective in mitigating class imbalance effects for 2D object detection.
The focal loss applies a dynamic weighting factor based on the usefulness of a given sample to re-weight the cross entropy, focusing on difficult-to-classify examples. \\
In particular, the focal loss proposes a modulating factor for a cross entropy loss:
\begin{equation}
    \mathcal{L}_\textrm{focal}(p_t) = -(1-p_t)^\gamma log(p_t),
\end{equation}
where $p_t$ is the point prediction probability for the respective target label and $\gamma\geq 0$ is focusing the modulating factor $(1-p_t)^\gamma$.

In practice, we did not see a direct improvement over cross entropy training by applying a focal loss directly, so we additionally re-balance the loss based on the class imbalance of the train set:
\begin{equation}
    FL(p_t) = - \alpha (1-p_t)^\gamma log(p_t), \quad\quad\quad \alpha_i = \frac{log(n_i)}{\sum_{j=1}^{N_\textrm{class}}log(n_j)}
\end{equation}

By explicitly considering category imbalances, we found this to provide improved performance over both a standard focal loss or direct category-balanced cross entropy (c.f. Sec~\ref{sec:results} for more analysis).

\subsection{Implementation Details}

During pre-training, we use a sparse 3D U-Net backbone for 3D feature encoding, implemented with the MinkowskiEngine~\cite{minkowski}.
We adapt the MinkUNet34 to output feature dimension maps of size $D=512$ to match the dimensionality of the pre-trained text encoding from CLIP~\cite{clip}. 
For additional details on optimization please refer our supplemental at Section 7. We follow a two stage training with pretraining and fine-tuning for both semantic and instance segmentation, where for all comparisons we use the same 3D backbone architecture.
\section{\OURS{} Benchmark}\label{sec:benchmark}

\begin{figure}[!ht]
    \centering
    \includegraphics[width=\textwidth,keepaspectratio]{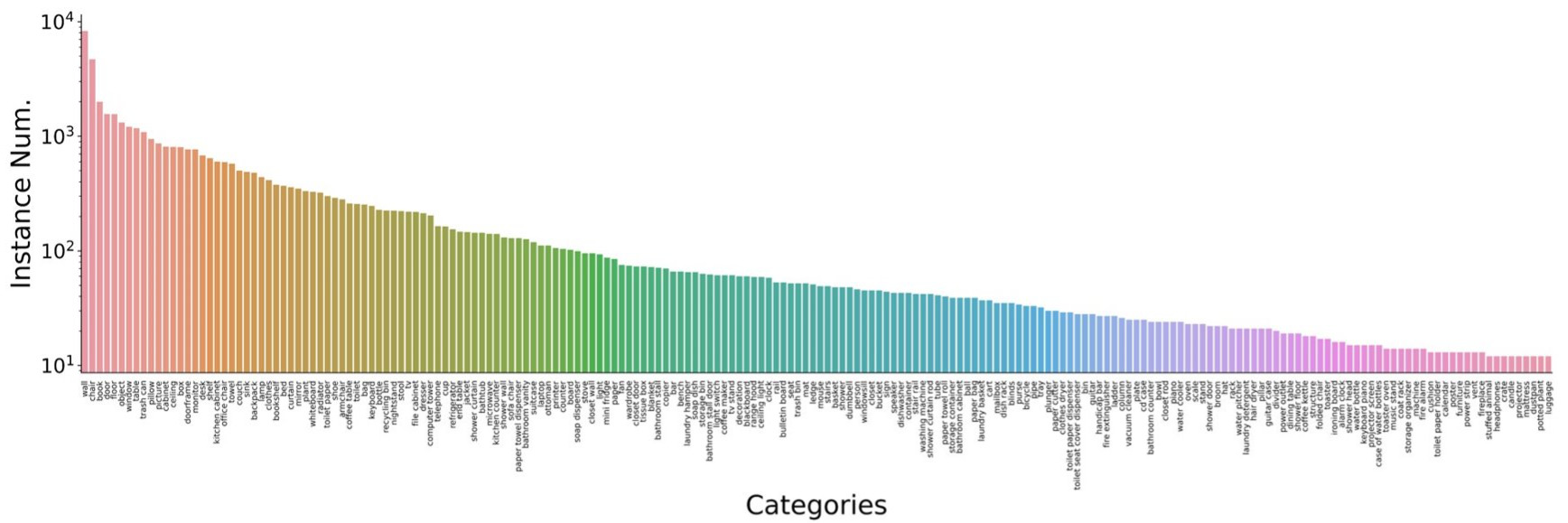}
    % \vspace{-0.3cm}
    \caption[]{Class category distribution for our \OURS{} Benchmark showing number of instances per category; note that the frequencies are given on log-scale and ordered by number of instances per category.} 
    \label{fig:histogram}
\end{figure}

The ScanNet Benchmark\footnote{\url{http://kaldir.vc.in.tum.de/scannet_benchmark/}} has provided an active online benchmark evaluation for 3D semantic segmentation, but only considers 20 class categories, which is insufficient to capture the diversity of many real-world environments.
We thus present the \OURS{} Benchmark for 3D semantic segmentation with \numlabels{} class categories, an order of magnitude more than previous.
We follow the original train/val/test split of ScanNet~\cite{scannet}, while training and evaluating over significantly more class categories.
Figure~\ref{fig:histogram} shows the class category distribution for \OURS{} over the number of annotated instances and the number of annotated surface points per category in the train set.

To obtain the \numlabels{} class categories, we considered the raw semantic label annotations provided by ScanNet~\cite{scannet}, which contains 607 raw categories.
After merging near-duplicate labels, this resulted in 550 unique semantic classes, from which we selected the \numlabels{}-most represented categories by the number of instances, forming \OURS{}.
The \numlabels{}-class selection enables enforcing a minimum of 10 samples from all categories.
%Note that these annotations contain semantic instance annotations for which instances are not considered during the evaluation of semantic segmentation.

In order to better understand performance under the natural class imbalance of the \OURS{} benchmark, we further split the \numlabels{} categories into sets of $66$, $68$ and $66$ categories, based on the frequency of number of labeled surface points in the train set: \textit{head}, \textit{common} and \textit{tail} respectively.
Evaluation over all categories as well as for the head, common, and tail splits enables a more precise understanding of segmentation performance.

\subsubsection{Limited Annotation Task.}
We additionally study semantic segmentation performance on \OURS{} in the limited annotation regime, as dense 3D annotations are expensive to acquire.
%In the limited annotation setting, we emulate annotations queried from annotators with one annotated point per object, similar to settings of weakly-supervised methods \cite{liu2021one} (as we do not consider zero-shot segmentation), and any additional points annotated based on farthest point sampling.
In the limited annotation setting, we emulate annotations queried from annotators with a randomly sampled annotated point per object, and any additional points annotated based on farthest point sampling, similar to settings of weakly-supervised methods \cite{liu2021one}.
We consider scenarios of (5\%, 10\%, 50\%) of annotated surface points provided, where all scene geometry is available (but unlabeled for surface points without annotations). 

\subsubsection{Instance Segmentation Task.}
In addition to 3D semantic segmentation, we also evaluate 3D instance segmentation on \OURS{}. We evaluate methods by mean Average Precision (mAP) at IoU of (25\%, 50\%) and averaged over all overlaps between $[50\%, 95\%]$ at 5\% steps, following the original \cite{scannet} benchmark.

\noindent
\subsubsection{Evaluation metrics.}
To evaluate semantic segmentation, we consider several evaluation metrics.
The primary evaluation metric is the category-level mean intersection-over-union (\textit{mIoU}) score as $tp / (tp + fp + fn)$, as a commonly adopted segmentation measure.
Additionally, we evaluate \textit{precision} as $tp/(tp + fp)$ and \textit{recall} as $tp/(tp + fn)$, to provide further insight towards over-prediction and under-prediction, respectively.
%
\begin{comment}
\textit{IoU}, \textit{precision}, and \textit{recall} are defined as follows:
%
\begin{equation*}
\begin{aligned}[c]
mIoU = \frac{tp}{tp + fp + fn}
\end{aligned}
\quad
\begin{aligned}[c]
Precision = \frac{tp}{tp + fp}
\end{aligned}
\quad
\begin{aligned}[c]
Recall = \frac{tp}{tp + fn}
\end{aligned}
\end{equation*}
%
\end{comment}
%
All evaluation metrics are measured across head, common, and tail splits as well as globally across all categories, in order to consider performance for more and less frequently seen class categories.
\section{Experiments}
\label{sec:results}

We evaluate our approach for language-grounded pre-training with state-of-the-art alternatives for 3D semantic segmentation on our \OURS{} benchmark.
For our method and all baselines, we use the same 80M parameter sparse 3D U-Net backbone implemented with MinkowskiNet~\cite{minkowski}.

\begin{table}[!ht]
	\centering
	\resizebox{\textwidth}{!}{\begin{tabular}{l|ccc|c?ccc|c?ccc|c}
		\multicolumn{1}{l}{\textbf{}} & \multicolumn{4}{c}{\textbf{mIoU}} & \multicolumn{4}{c}{\textbf{Precision}} & \multicolumn{4}{c}{\textbf{Recall}} \\
		\multicolumn{1}{l}{\textbf{}} & Head  & Common & Tail  & All   & Head  & Common & Tail  & All   & Head  & Common & Tail  & All    \\ \hline \hline
		Scratch             & 48.29 & 19.08  & 7.86 & 25.02 & 68.81 & 66.29  & 39.88 & 58.32 & 60.45 & 25.50  & 15.06 & 33.67 \\ \hline
		Ins. samp.         & 48.46 & 18.97  & 9.22 & 25.49 & 70.04 &  62.98 & 49.41 & 60.81 & 59.64 & 24.66  &  19.25 & 34.52 \\ \hline
		C-Focal        &   48.10    &   20.28     &   9.38    &   25.86    &   68.10    &    65.64    &   47.43    &  60.39     &   60.08    &     26.28   &  19.14     &   35.48   \\ \hline
		SupCon~\cite{supcontrast}    & 48.55 & 19.17  & 10.34 & 26.02 & 69.52 & 65.42  & 40.62 & 58.52 & 60.27 & 26.28  & 19.14  & 35.23 \\ \hline
		CSC~\cite{scene_contrast}    & 49.43 & 19.52  & 10.28 & 26.41 & 70.00 & 67.75  & 40.78 & 59.51 & 61.01 & 25.75  & 17.62  & 34.79 \\ \hline  
		Ours (CLIP only)            & 50.39 & \textbf{22.84}  & 10.10 & 27.73 & 71.64 & \textbf{69.72}  & 44.47 & 61.94 & 62.20 & \textbf{29.37}  & 17.35 & 36.16\\ \hline
		\textbf{Ours}             & \textbf{51.51} & 22.68  & \textbf{12.41} & \textbf{28.87} & \textbf{72.72} & 66.69  & \textbf{58.30} & \textbf{65.90} & \textbf{62.50} & 29.09  & \textbf{26.61} & \textbf{39.40} \\ \hline 
		\end{tabular}}
	\caption{Comparison to state of the art on \OURS{}. Our language-grounded 3D feature learning enables improved performance across frequent and infrequently seen categories in comparison with pure data augmentation or loss balancing techniques as well as state-of-the-art 3D pre-training. 
	Our approach achieves over 5\% mIoU performance over training from scratch, more than double the performance improvement of CSC~\cite{scene_contrast}.}
	\label{tab:large_model_results}
\end{table}

\begin{figure}[!t]
    \centering
    \includegraphics[width=\linewidth]{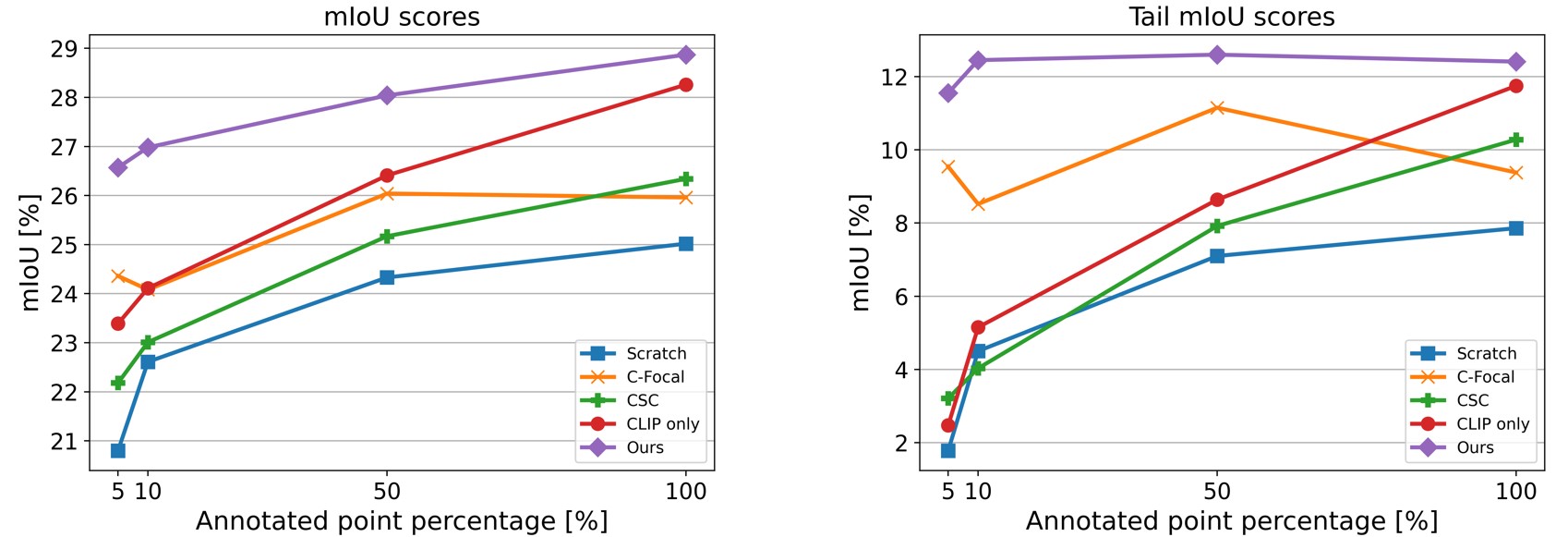}
    % \vspace{-0.5cm}
    \caption{
    3D semantic segmentation under varying amounts of limited annotations.
    Even when considering only a small number of annotated surface points for our supervised language-guided 3D pre-training, our approach improves notably over the state-of-the-art 3D pre-training of CSC~\cite{scene_contrast}. }
    \label{fig:limitedann}
\end{figure}

\smallskip
\noindent \textbf{Comparison to the state of the art.}
We compare with a state-of-the-art  pre-training approaches Contrastive Scene Contexts (CSC) \cite{scene_contrast} and Supervised Contrastive Learning (SupCon) \cite{supcontrast}, along with our instance-based data balancing and focal loss~\cite{focalloss} training in Table~\ref{tab:large_model_results}.
For CSC, we use the same pre-training experimental setup as proposed by the authors for our 3D backbone. 
For SupCon, we sample 5 positive and 5 negative candidates from the training scene for each source point and train it for 300 epochs with the same optimization parameters as our method.
Our instance sampling, as well as focal loss, individually help to improve performance, particularly for lesser-seen class categories.
Additionally, all pre-training approaches improve performance over training from scratch, while our language-grounded feature learning enables more effective semantic reasoning with consistent improvements over baselines and across common and rarely seen class categories.

\begin{figure}
    \centering
    \includegraphics[width=\textwidth]{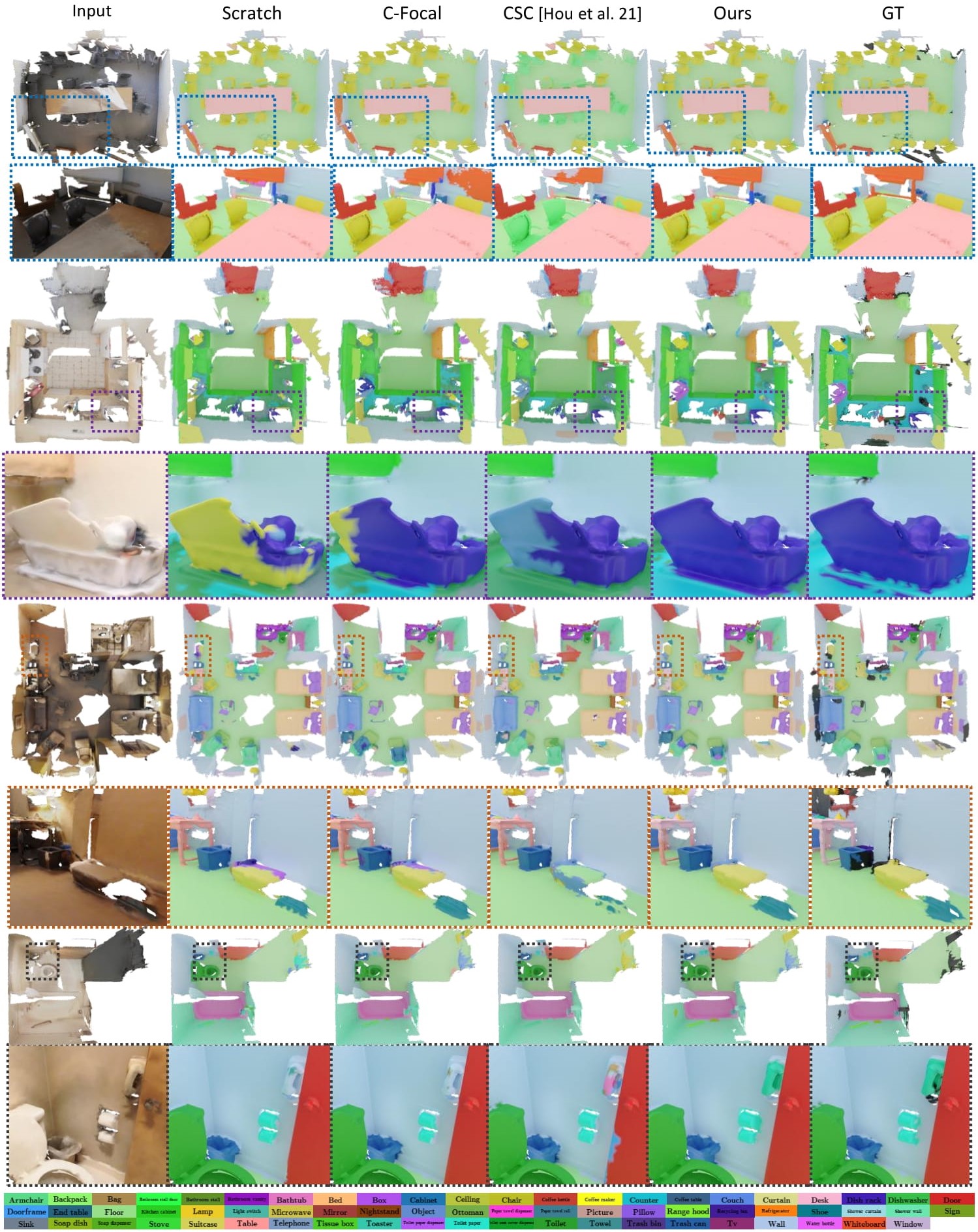}
    %\vspace{-0.5cm}
    \caption{Qualitative semantic segmentation results on ScanNet~\cite{scannet} scenes. 
    In comparison to training from scratch, class-balance focal loss, and the 3D pre-training of CSC~\cite{scene_contrast}, our language-grounded 3D feature learning enables more consistent and accurate semantic segmentation, even for challenging less frequently seen class categories (e.g., \textit{``dish rack"} in row 4, \textit{``telephone"} in the last row).}
    \label{fig:methods_comparison}
\end{figure}

\smallskip
\noindent \textbf{Limited annotation semantic segmentation.}
As data annotations remain expensive to acquire, we additionally evaluate our approach in comparison with state of the art in the limited annotation scenario of our \OURS{} Benchmark described in Sec.~\ref{sec:benchmark}.
Figure~\ref{fig:limitedann} shows performance over varying amounts of labeled annotation data available (5\%, 10\%, 50\%, 100\%).
Note that since our pre-training leverages text labels to guide pre-training, we only pre-train with the available annotations, whereas CSC is pre-trained with all geometric data available for the train scenes and fine-tuned with the limited annotation data.
Our approach enables more robust semantic segmentation on this challenging benchmark, consistently improving and recovering the performance of training from scratch with only 5\% of the annotations. Moreover, in the very low annotation regime, we see significant improvements on tail categories, with an increase of +8 mIoU from the state-of-the-art 3D pre-training of CSC with 5\% of annotations available.

\smallskip
\noindent \textbf{How much does a class-balanced focal loss help?}
We evaluate the effect of our class-balanced  focal loss~\cite{focalloss} variant (\emph{C-Focal}) in Table~\ref{tab:large_model_results}, which helps to improve performance over training from scratch with a standard cross entropy loss.
Additionally, we see a consistent improvement with a smaller 3D backbone model in Table 3 in supplementary material, particularly for tail categories. We note that the class-balanced focal loss improves notably over both the original focal loss formulation (both using $\gamma = 2$), as well as a class-balanced cross entropy.

\smallskip
\noindent \textbf{What is the impact of data balancing with instance sampling?}
We additionally evaluate the effect of applying data balancing by our instance sampling during training in Table~\ref{tab:large_model_results} (\emph{Ins. samp}) as well as for a smaller 3D backbone in supplemental Table 3. 
We find that this instance sampling consistently provides a small improvement in performance across common and rare class categories.

\smallskip
\noindent \textbf{What is the effect of our language-grounded pre-training?}
Table~\ref{tab:large_model_results} shows that our language-grounded pretraining to text-based CLIP~\cite{clip} embeddings without focal loss or instance sampling already improves over all baselines.
Our full approach with focal loss and instance sampling in addition to text-anchored pre-training enables consistent, effective improvement in comparison to alternative approaches.

\smallskip
\noindent \textbf{3D instance segmentation task.}
In addition to 3D semantic segmentation, we also analyze a 3D instance segmentation task in Table \ref{tab:instseg_results}, showing that our approach generalizes across multiple downstream tasks with consistent performance improvement.
We use the same pre-trained 3D backbones and fine-tune them for instance segmentation by predicting an offset vector for every scene point as a voting mechanism together with the semantic labels. These directional distance vectors are optimized during train time, while the clustering of the instances is calculated only at test time. For the task and clustering algorithm, we adopt the paradigms of  \cite{jiang2020pointgroup,scene_contrast} to our \OURS{} benchmark. For this task, we train our models with a batch size of $8$ for $300$ epochs and momentum SGD optimizer with the same parameters as in the semantic segmentation experiments, except for a smaller starting learning rate of $0.02$.
\begin{table}
\centering
\resizebox{0.5\textwidth}{!}{\begin{tabular}{lc?c?c}
  &
  \textbf{Precision} &
  \textbf{mIoU} &
  \textbf{mAP@0.5} \\ \hline \hline
\multicolumn{1}{p{2cm}|}{Scratch} &
  \multicolumn{1}{>{\centering\arraybackslash}p{2cm}?}{61.04} &
  \multicolumn{1}{>{\centering\arraybackslash}p{2cm}?}{25.37} &
  \multicolumn{1}{>{\centering\arraybackslash}p{2cm}|}{24.47} \\ \hline
\multicolumn{1}{l|}{CSC~\cite{scene_contrast}} &
  \multicolumn{1}{c?}{63.13} &
  \multicolumn{1}{c?}{25.92} &
  \multicolumn{1}{c|}{25.24} \\ \hline
\multicolumn{1}{l|}{CLIP only} &
  \multicolumn{1}{c?}{64.24} &
  \multicolumn{1}{c?}{27.58} &
  \multicolumn{1}{c|}{\textbf{27.91}} \\ \hline \hline
\multicolumn{1}{l|}{Ours} &
  \multicolumn{1}{c?}{\textbf{65.32}} &
  \multicolumn{1}{c?}{\textbf{27.72}} &
  \multicolumn{1}{c|}{26.09} \\ \hline
\end{tabular}}
\caption{3D instance segmentation, in comparison with training from scratch and state-of-the-art 3D pre-training approach CSC~\cite{scene_contrast}. Our language-grounded pre-training improves over both baselines.}
\label{tab:instseg_results}
\end{table}

\begin{figure}[!ht]
    \centering
    \begin{subfigure}[c]{.24\linewidth}
        \includegraphics[width=\linewidth]{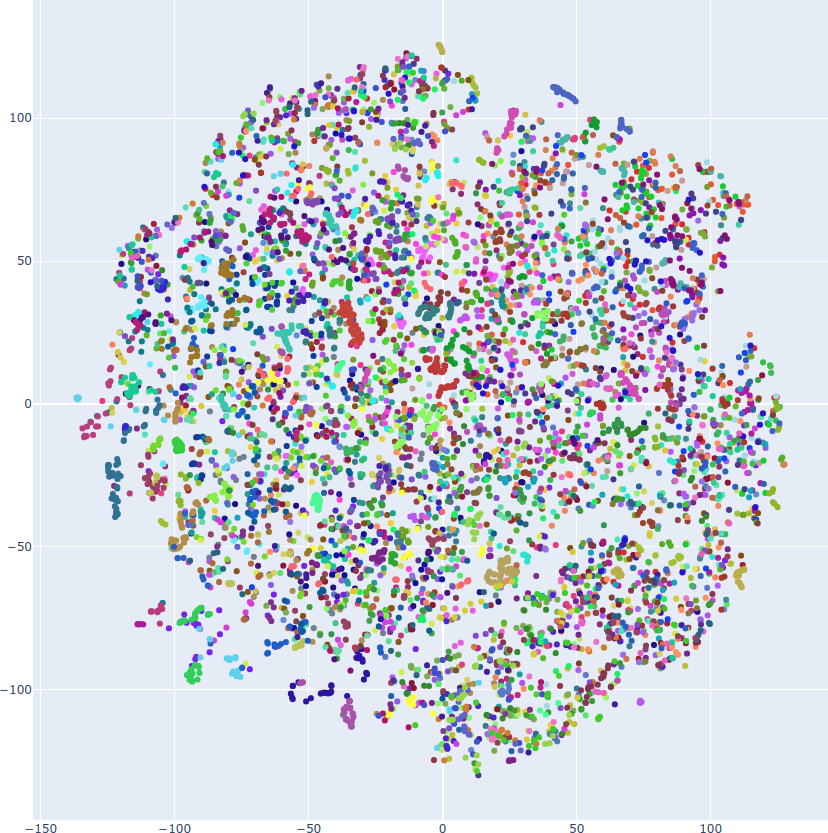}
        \caption{CSC~\cite{scene_contrast}}
    \end{subfigure}
    \begin{subfigure}[c]{.24\linewidth}
        \includegraphics[width=\linewidth]{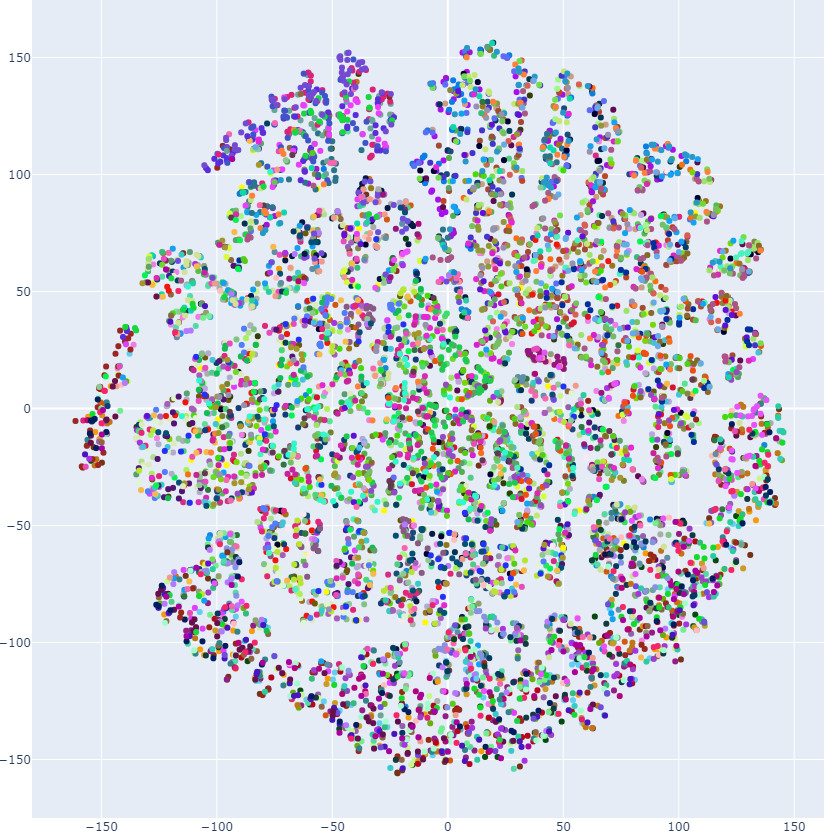}
        \caption{SupCon~\cite{supcontrast}}
    \end{subfigure}
    \begin{subfigure}[c]{.24\linewidth}
        \includegraphics[width=\linewidth]{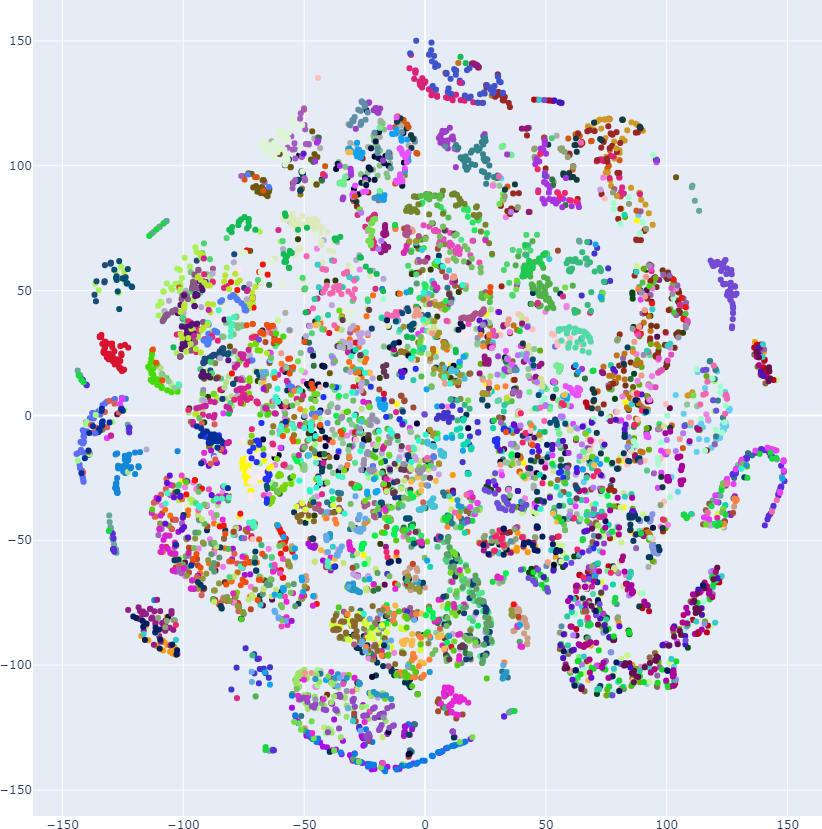}
        \caption{Ours (only pos.)}
    \end{subfigure}
    \begin{subfigure}[c]{.24\linewidth}
        \includegraphics[width=\linewidth]{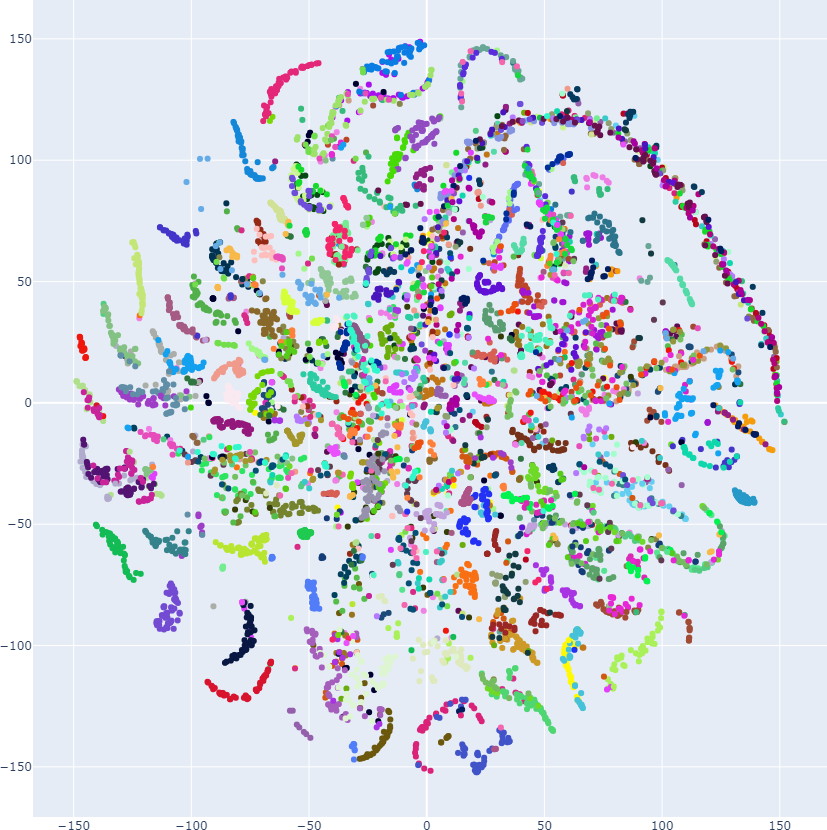}
        \caption{Ours}
    \end{subfigure}
    %
    %
    % \vspace{-0.2cm}
    \caption{We show a comparison with the representation learned by CSC \cite{scene_contrast}, SupCon~\cite{supcontrast}, as well as our approach when training with only positive samples. Our full language-grounded pre-training results in a more structured feature representation space with improved semantic segmentation performance.}
    \label{fig:representation_spaces}
\end{figure}

%\smallskip
\noindent \textbf{Learned feature representation space.}
We analyze the pre-trained representation spaces by visualizing a t-SNE projection of the learned features in Figure~\ref{fig:representation_spaces}.
By anchoring 3D feature learning to a richly-structured text embedding space, we can learn a more structured 3D feature representation space. % to support improved 3D semantic segmentation performance.

\paragraph{Limitations and Future Work.}
We believe our language-grounded 3D feature learning provides a promising step towards more robust and general 3D scene understanding, though several important limitations remain.
It is often the case that infrequently observed objects are small and their geometric resolution is limited, so while tail category performance has improved using only geometric input, there is still much room for improvement.
In particular, we note that color image observations could provide significantly higher resolution signals to explore for more accurate tail category recognition.
Additionally, text encodings are used to anchor learned 3D feature representations, but currently, only the semantic labels of each object are considered, whereas text caption or object attribute descriptions could potentially provide a richer signal.

\section{Conclusion}

We have presented \OURS{}, a new benchmark for 3D semantic segmentation with an order of magnitude more class categories, along with a new approach for language-grounded pre-training to address 3D semantic feature learning under imbalanced and limited data.
Our approach demonstrates robust feature learning by anchoring learned features to richly-structured CLIP text embeddings, demonstrating consistent improvements over strong baselines on our challenging \OURS{} Benchmark and under limited annotation scenarios.
We believe that this makes an important step towards 3D semantic scene understanding in the wild, and forms a basis for future multi-modal exploration for a variety of 3D perception tasks.

\section*{Acknowledgements}
This project is funded by the Bavarian State Ministry of Science and the Arts and coordinated by the Bavarian Research Institute for Digital Transformation (bidt).

%\clearpage
\bibliographystyle{splncs04}
\bibliography{main}

\clearpage
\section*{Appendix}
In this supplemental material, we show additional results on the downstream task of 3D instance segmentation and the extended benchmark of \OURS{} in Sec.~\ref{sec:inst_segmentation}. We additionally provide further ablation analysis in Sec.~\ref{sec:distance_metric} on the effect of the pretrained language model and distance metric used in the contrastive objective of the point-to-language features. Finally, we present a break-down of per-class IoU scores in Sec.~\ref{sec:breakdown}.

\section{Additional Ablations}\label{sec:distance_metric}

\paragraph{Generalization across backbone sizes.}
In Table~\ref{tab:small_model_results}, we evaluate our approach and baselines using a smaller 20M parameter 3D backbone model. 
In this scenario, we do not map 3D features directly to their $512$-dimensional text embeddings from CLIP but a $96$-dimensional projection (obtained by PCA). 
We see consistent improvements from training from scratch as well as over state of the art while using a smaller-backbone architecture.
\begin{table}[!ht]
	\centering
	\resizebox{\textwidth}{!}{\begin{tabular}{c|ccc|c?ccc|c?ccc|c}
		\multicolumn{1}{l}{\textbf{}} & \multicolumn{4}{c}{\textbf{mIoU}} & \multicolumn{4}{c}{\textbf{Precision}} & \multicolumn{4}{c}{\textbf{Recall}} \\
		& Head  & Common & Tail & All   & Head  & Common & Tail  & All   & Head  & Common & Tail  & All   \\ \hline \hline
		Scratch         & 45.50 & 13.64  & 3.41 & 20.78 & 66.89 & 55.69  & 23.30 & 48.62 & 57.59 & 19.12  & 5.83  & 27.78 \\ \hline
		Ins. samp.      & 47.70  & 14.40  & 5.33 & 22.48 & 69.01 & 58.71  & 36.33 & 54.68 & 59.95 & 51.75  & 13.07  & 31.59 \\ \hline
		Weighted CE          & 46.99 & 16.72  & 6.24 & 23.25 & 67.33 & 62.15  & 35.62 & 55.03 & 59.76 & 23.69 & 12.58  & 31.80 \\ \hline
		Focal~\cite{focalloss}         & 48.31 & 15.43  & 4.61 & 22.78 & 66.92 & 62.68  & 24.95 & 51.52 & 58.22  & 22.86  & 9.87   & 30.31 \\ \hline
		C-Focal                & 46.47 & 17.15  & \textbf{9.45} & 24.29 & 66.27 & 60.24  & 38.25 & 54.92 & 59.47 & 23.25  & 16.27 & 32.90 \\ \hline
		CSC~\cite{scene_contrast}                & 48.51 & 16.29  & 5.88 & 23.49 & \textbf{70.99} & 62.16  & 29.35 & 54.16 & 60.47 & 21.31  & 9.46  & 30.19 \\ \hline
		\textbf{Ours}                 & \textbf{48.88} &\textbf{ 19.97}  & 9.03 & \textbf{25.93} & 68.77 & \textbf{65.36}  & \textbf{42.33} & \textbf{58.82} & \textbf{61.34} & \textbf{27.75}  & \textbf{19.11} & \textbf{35.49} \\ \hline
		\end{tabular}}
	\caption{Generalization across backbone sizes: 3D semantic segmentation with a 20M parameter 3D U-Net backbone on \OURS{}. Our approach maintains consistent improvements over state of the art with this smaller 3D backbone.}
	\label{tab:small_model_results}
\end{table}

\paragraph{Comparison with point-based baselines}
As an additional ablation we compare out method with point-based state-of-the-art segmentation models capable of processing complete ScanNet scenes at once. We choose RandLa-Net \cite{RandLA-Net} and SCF-Net \cite{Fan_2021_CVPR_SCFNet} as baselines and use the official authors implementation and hyperparameters for both methods. Our method outperforms both approaches with our language-guided pretraining on this challenging large-vocabulary task. The performance evaluated on ScanNet200 can be seen in Table \ref{tab:point_based}.

\begin{table}[!h]
\centering
\begin{tabular}{lccc|c}
                                & Head & Common & Tail & Mean                  \\ \hline \hline
\multicolumn{1}{l|}{RandLA-Net} &    35.35    &  5.15      &   0.87    & \multicolumn{1}{c|}{15.06} \\ \hline
\multicolumn{1}{l|}{SCF-Net}    &     35.99   &    5.97    & 0.38   & \multicolumn{1}{c|}{15.45} \\ \hline
\multicolumn{1}{l|}{\textbf{Ours}}       &    \textbf{51.51}    &  \textbf{22.68}    &    \textbf{12.41}      & \multicolumn{1}{c|}{\textbf{28.87}} \\ \hline
\end{tabular}
%\vspace{-0.2cm}
\caption{Comparison with point-based RandLA-Net and SCF-Net on ScanNet200 semantic segmentation (mIoU). }
\label{tab:point_based}
\end{table}

\paragraph{Effect of contrastive distance metric.}
Table~\ref{tab:language_ablation} additionally considers alternative distance metrics of $\ell_1$ and $\ell_2$ in comparison with our used cosine distance metric.
The $\ell_1$ and $\ell_2$ distances were more challenging to optimize to align to corresponding text embeddings, with cosine distance producing the best performance.

\paragraph{Effect of the pre-trained language model.}
In Table~\ref{tab:language_ablation}, we consider alternative language models to CLIP~\cite{clip}; both BERT~\cite{bert} and GPT2~\cite{gpt2}  are also popular language models trained on large amounts of text data, rather than the image-text training of CLIP.

For text encodings, we use the BERT variant \textit{bert\_uncased\_L-8\_H-512\_A-8} from \cite{turc2019well}, and project the $768$-dimensional GPT2 encodings from the small GPT2 model to $512$ dimensions by PCA.
We find the rich embedding structure from the multi-modal nature of CLIP produces the best results.

\begin{table}[!h]
	\centering
	\resizebox{\textwidth}{!}{\begin{tabular}{c|ccc|c?ccc|c?ccc|c}
		\multicolumn{1}{l}{\textbf{}} & \multicolumn{4}{c}{\textbf{mIoU}} & \multicolumn{4}{c}{\textbf{Precision}} & \multicolumn{4}{c}{\textbf{Recall}} \\
		& Head  & Common & Tail & All   & Head  & Common & Tail  & All   & Head  & Common & Tail  & All   \\ \hline \hline
		Scratch         & 48.29 & 19.08  & 7.86 & 25.02 & 68.81 & 66.29  & 39.88 & 33.67 & 60.45 & 25.50  & 15.06 & 33.67 \\ \hline
		GPT2 \cite{gpt2}         & 45.70 & 19.07 & 9.73 & 24.78 & 69.42 & 66.18 & 48.76 &  & 56.86 & 25.132 & 16.75 & 32.41 \\ \hline
		BERT \cite{bert}      & 47.70 & 18.19 & 9.16 & 24.95 & 70.48 & 64.09 & 42.16 &  & 58.24 & 24.01 & 16.19 & 32.65 \\ \hline \hline
		BERT $\ell_2$         & 41.16 & 14.02 & 8.89 & 21.28 & 67.82 & 61.65 & 39.38 & & 53.90 & 19.86 & 13.72 & 20.77 \\ \hline
		CLIP \cite{clip} $\ell_1$        & 39.28 & 10.26 & 2.80 & 17.38 & 64.52 & 57.64 & 27.57 & & 48.81 & 14.09 & 3.95 & 22.29\\ \hline
		CLIP $\ell_2$               & 43.48 & 16.97 & 8.87 & 23.04 & 67.15 & 64.64 & 41.54 & & 54.50 & 22.07 & 14.77 & 29.92\\ \hline \hline
		CLIP  & \textbf{50.39} & \textbf{22.84}  & \textbf{10.10} & \textbf{27.73} & \textbf{71.64} & \textbf{69.72}  & \textbf{44.47} & \textbf{42.67} & \textbf{62.20} & \textbf{29.37}  & \textbf{17.35} & \textbf{36.16}\\ \hline
		\end{tabular}}
	\caption{Ablation study on different language models for generating the text anchors during the pre-training stage. We show that while the model benefited from pretraining guided by all language models, CLIP was found to be the most suitable for this task. We also show that more rigid loss distance metrics such as $l1$ or $l2$ can even significantly hinder the performance.}
	\label{tab:language_ablation}
\end{table}

\section{Implementation Details}\label{sec_sup:inst_sampling}
\subsubsection{Training parameters}

In the pretraining stage, we use a momentum SGD optimizer with batch size 8 and an initial learning rate of 0.05, decayed by a factor of 0.3 at epochs $150$ and $250$, a momentum of $0.9$, and train for $400$ epochs until convergence. 
We additionally use $\lambda = 1$, $t_{pos} =  0$, $t_{neg} = 0.6$ and $N_i = 3$ uniformly sampled from all ScanNet200 categories.

We then fine-tune the pre-trained 3D backbone for 3D semantic segmentation.
We optimize with the same momentum SGD  and batch size, with an initial learning rate of 0.05, decayed by a factor of 0.3, and train for 170 epochs until convergence.
For the instance sampling, we sampled from the 66 classes least frequently represented in the training set surface point annotations.

\subsubsection{Instance Sampling}
For the instance sampling, we randomly sample from the 66 tail classes, and select them by probability computed from the inverse log frequencies of the train set histogram. 
Object centers are  placed in the scene by randomly sampling $(x,y)$ locations in the scene, with $z$ determined by the max height  of the scene geometry at $(x,y)$ (such that the object sits on the scene geometry).  Objects are then inserted with a random orientation around the up ($z$) axis.
Any object insertion that induces a bounding box collision with scene objects or previously inserted objects are discarded.
Placement determined by object center on the support plane encourages placement of instances with sufficient physical support in the scene.
In combination with color and geometry augmentation, this helps to learn more robust features in our large-vocabulary setting. 

We also note here that while we do not explicitly address lighting effects during instance augmentation, a minor appearance difference is noticeable on the original and sampled parts of the scene, we still achieve a clear effect with this augmentation technique. 
We hypothesize that at the 2cm resolution which our method (and state-of-the-art methods) use, lighting inconsistencies have a limited-to-negligible effect. In addition, we use random color jittering, which reduces the chance of learning from erroneous signal, and observe that the advantage of this geometric augmentation provides notably more benefit than color. 
We provide evidence of our reasoning with an experiment to train the same baseline sparse 20M parameter UNet model with and without voxel color signal, and observed that the final performance differs only in 1\% mIoU.

\section{3D Instance Segmentation Results}\label{sec:inst_segmentation}
\begin{figure}[!htb]
    \centering
    \includegraphics[width=\textwidth]{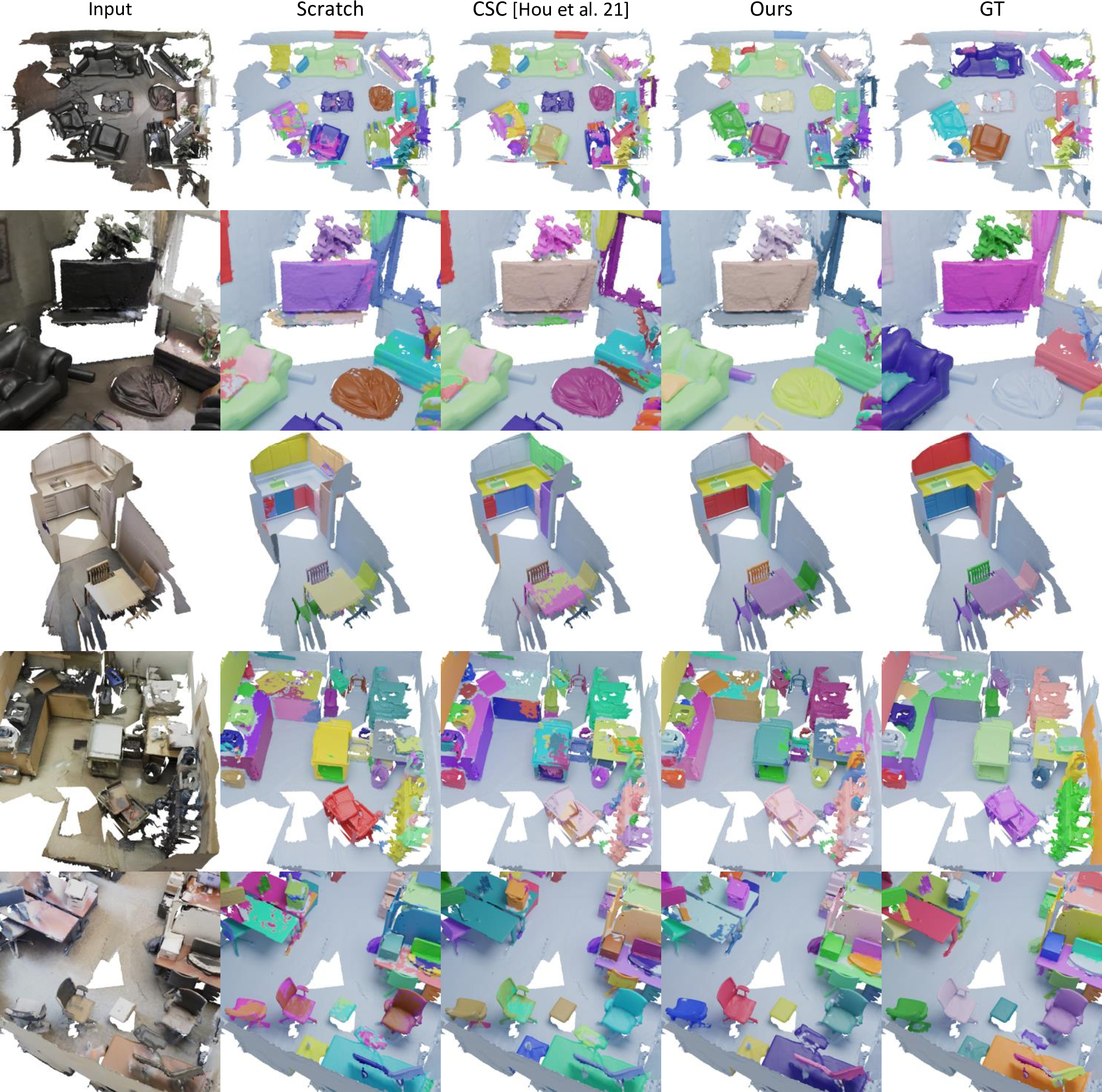}
    %\vspace{-0.5cm}
    \caption{Qualitative results for 3D semantic instance segmentation results on ScanNet~\cite{scannet} scenes.
    Our language-grounded pretraining together with class-balanced losses can also effectively improve performance in object recognition.}
    \label{fig:instseg_results}
\end{figure}

Figure \ref{fig:instseg_results} shows qualitative visualizations for the downstream task of 3D instance segmentation in comparison to training from scratch and CSC~\cite{scene_contrast}.

\section{Breakdown of Class IoU scores}\label{sec:breakdown}

Table~\ref{tab:class_iou_2} shows the per-category IoU scores for 3D semantic segmentation on \OURS{}.

\begin{table}[ht]
\centering
\includegraphics[width=\linewidth,trim={2cm 0 2cm 0},clip]{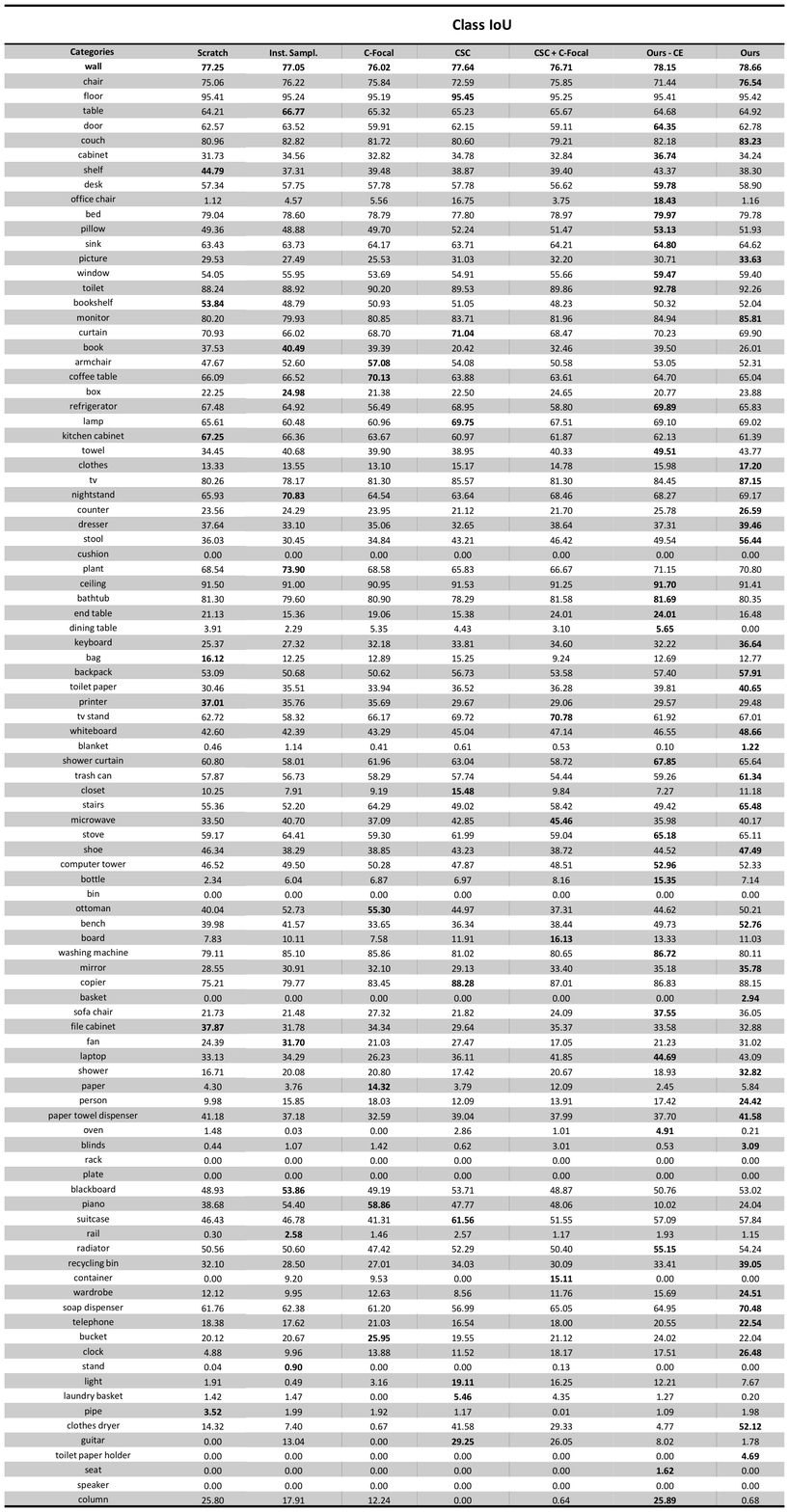}
\label{tab:class_iou_1}
\end{table}

\begin{table}[ht]
\centering
\includegraphics[width=\linewidth,trim={2cm 0 2cm  0},clip]{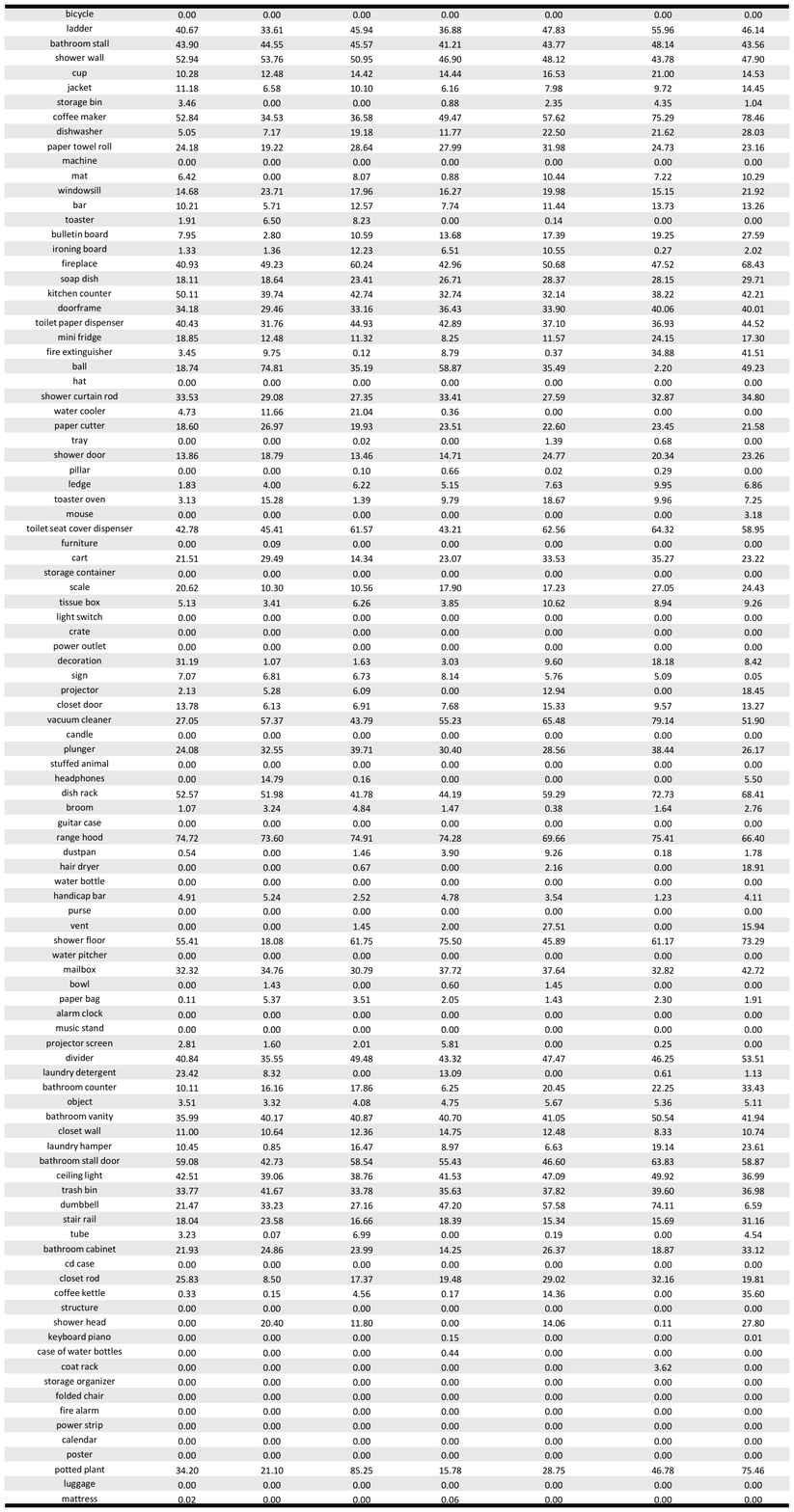}
\caption{Class IoU scores on the \OURS{} benchmark of our proposed method, and compared with other state-of-the-art approaches.}
\label{tab:class_iou_2}
\end{table}

\end{document}